\documentclass[oribibl,envcountsame]{llncs}  

\usepackage{epsfig}
\usepackage{tikz}
\newcommand{\ignoreText}[1]{}
\usepackage{multirow}
\usepackage{rotating}
\usepackage{array}
\usepackage{amsmath, amssymb}
\usepackage{subfigure}

\usetikzlibrary{shapes,snakes}

\newcommand{\hpopc}{\ensuremath{\mathsf{HPO_{PC}}}}
\newcommand{\hpopcS}{\ensuremath{\mathsf{HPO_{PC}}~}}
\newcommand{\hpomerged}{\ensuremath{\mathsf{HPO_{ALL}}}}
\newcommand{\hpomergedS}{\ensuremath{\mathsf{HPO_{ALL}}~}}
\newcommand{\dex}[1]{\exists#1.}

\newcommand{\dfa}[1]{\forall#1.}
\newcommand{\deq}{\equiv}

\title{First steps in the logic-based assessment of post-composed phenotypic
descriptions}

\author{%
  E. Jim\'enez-Ruiz\inst{1,3} \and   
  B. Cuenca Grau\inst{2}      \and
  R. Berlanga\inst{1}  		\and 
  Dietrich Rebholz-Schuhmann\inst{3}
}
\institute{%
  Universitat Jaume I de Castell\'o, Spain 
  \{ejimenez,berlanga\}@uji.es                                     \and%
 University of Oxford, UK  
  \{berg\}@comlab.ox.ac.uk   \and
 European Bioinformatics Institute, Cambridge, UK
 \{jimenez,rebholz\}@ebi.ac.uk
}

\begin{document}

\maketitle

\begin{abstract}
In this paper we present a preliminary logic-based evaluation of the
integration of post-composed phenotypic descriptions with domain ontologies.
The evaluation has been performed using a description logic reasoner together
with scalable techniques: ontology modularization and approximations of the
logical difference between ontologies.
\end{abstract}

\section{Introduction}

A phenotype is defined as a basic observable
characteristic of an organism. 
Thus, a set of phenotypic descriptions may involve different domains and
granularities ranging from molecular to organism level. 


%
Phenotypic descriptions have been recently described by means of terminological
resources, with the Human Phenotype Ontology (HPO) \cite{HPO2010} being a prominent example. The
HPO ontology represents a so-called pre-composed description: it does not provide explicit
links between the phenotypic descriptions (e.g. increased calcium concentration
in blood) and the relevant entities associated to it, such as the chemical element involved
(``calcium''), the way in which it is involved (``increased concentration'') and where it appears
(``blood'').
Post-composed phenotypic
descriptions intend to provide a more formal representation to interoperate with involved entities \cite{Lussier2004}
and to allow more powerful reasoning. Nevertheless, the formal representation 
of phenotypic descriptions is still a challenge
\cite{MungallPheno2010,Hoehndorf2010} owing to the complex nature of some
phenotypes and the lack of consensus among clinicians to describe them in a
standard way. 

Mungall et al. \cite{MungallPheno2010} and Hoehndorf et al. \cite{Hoehndorf2010}
have recently proposed automatic and semi-automatic methods to transform
pre-composed phenotypic descriptions into a \textit{description logic} (DL)
based post-composed representation linked to domain ontologies.
The integration of domain ontologies with post-composed phenotypic
descriptions presents new challenges since most of the involved ontologies
are developed independently and may perform a different conceptualization for
the same entities. Therefore, this integration may not always lead to the expected
and proper logical consequences \cite{ESWC2009,umlsassessment10}. 
In this paper we present first steps towards the logic-based assessment of the
integration of phenotypic descriptions with domain ontologies.

\section{Method and preliminary results}

\begin{figure*}[t]
\centering{
	\includegraphics[width=0.75\textwidth]{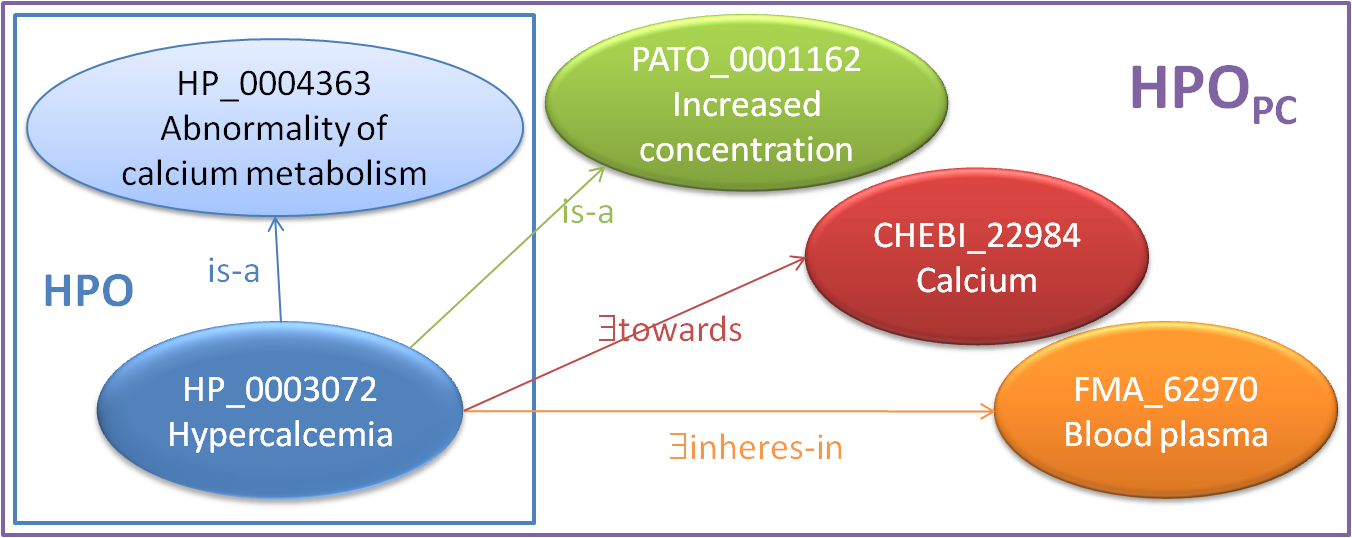}
    \caption{An excerpt from the post-composed phenotypic descriptions of \hpopc}
\label{fig:hpo_links}
}
\end{figure*}

Our experiments have been based on a post-composed version (from now on
\hpopc) of the HPO ontology\footnote{Available from
\url{http://bioonto.de/obo2owl/hpo-in-owl.owl}} applying the method
from \cite{MungallPheno2010}. The HPO ontology only provides a classification of
pre-composed phenotypic descriptions (e.g. see left hand side of Figure
\ref{fig:hpo_links}), whereas \hpopcS also provides explicit links
to relevant domain entities (see right hand side of Figure
\ref{fig:hpo_links}).
\hpopcS contains 11382 entities and uses \textit{external} concepts from
different domain ontologies, including PATO \cite{PATO2004} (264 concepts), Cell Ontology (12
conc.), GO (96 conc.), FMA \cite{FMA2004} (812 conc.),
CHEBI (33 conc.), and other OBO foundry ontologies \cite{obo2007}.

A DL reasoner may be used to reclassify HPO concepts, according to
the knowledge of \hpopcS and linked ontologies, and get new interesting
knowledge. However, as stated in \cite{MungallPheno2010}, reasoning with
\hpopcS and all linked ontologies is time consuming. 
To smooth this limitation, we have extracted a locality-based module
\cite{GrauJAIR08} for each set of referenced external entities. For example, the module for
FMA contains 2044 concepts, 
which is much easier to reason with than the whole FMA (around 80000 concepts).
Thus, we have built 
\hpomerged\footnote{We
have converted the OBO ontologies to OWL using the OWLDEF method
\cite{owldeff2010}, and we have normalized the involved concept and property
URIs} 
by merging \hpopcS with the corresponding modules from the referenced
ontologies. The classification of \hpomergedS using HermiT \cite{hermit2009} takes around
45 seconds in a 2Gb laptop.

New subsumption relationships between HPO concepts may represent
both desired new knowledge and unintended consequences.
In order to evaluate the new logical consequences hold in \hpomergedS we have
borrowed the notion of logical difference from \cite{KonevWW08}. The logical
difference between two ontologies contains the set of consequences that
are inferred in one of the ontologies but not in the other. Unfortunately, there is
no algorithm for computing the logical difference in expressive DLs. Moreover,
the number of inferences in the difference may be infinite.
Thus, we have reused the approximations of the logical difference presented in
previous work \cite{ESWC2009}, where inferences are one of the following simple
kinds of axiom: \emph{(i)} $A \sqsubseteq B$, \emph{(ii)} $A \sqsubseteq \neg
B$, \emph{(iii)}  $A \sqsubseteq \dex{R}B$, \emph{(iv)} $A \sqsubseteq
\dfa{R}B$, and \emph{v)} $R \sqsubseteq S$ ($A,B$ are atomic concepts,
including $\top,\bot$, and $R,S$ atomic roles).

\begin{figure*}[t]
\centering{
    \includegraphics[width=0.45\textwidth]{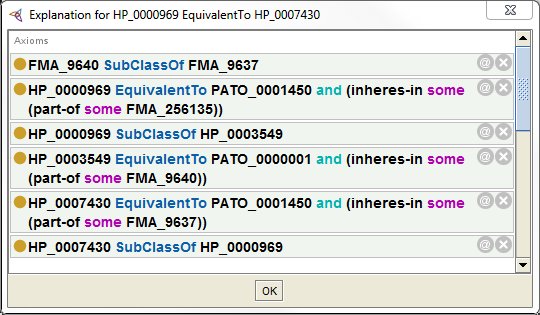}
    \includegraphics[width=0.45\textwidth]{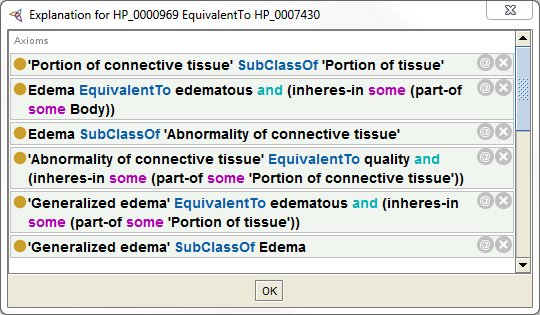}
\caption{Explanation for new equivalence between concepts $HP\_0000969$
($Edema$) and $HP\_0007430$ ($Generalized~edema$). With concept IDs (left) and
concept names (right).}
\label{fig:edema_exp}
}
\end{figure*}

The logical difference between \hpomergedS and \hpopc,
affecting only HPO concepts, contains 759 new subsumption
relationships (inferences of type \emph{(i)}).
The integration leads indeed to a reclassification of HPO
concepts. 
For example,
\hpomergedS infers the probably non-intended consequence 
$Generalized~edema \deq Edema$ 
which was not hold in \hpopc. As shown in the
Prot\'eg\'e-like explanation from Figure \ref{fig:edema_exp} the new knowledge
from FMA leads to this new consequence.

\begin{figure*}[tb]
\centering{
    \includegraphics[width=0.45\textwidth]{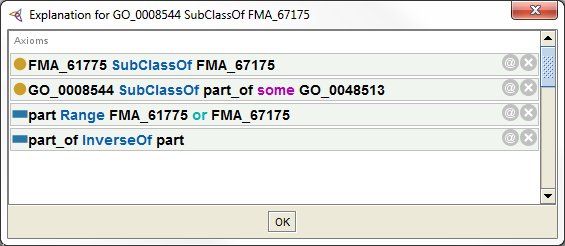}
    \includegraphics[width=0.45\textwidth]{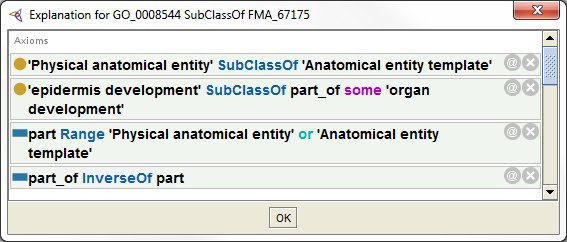}
\caption{Explanation for new subsumption relationship between concepts
$GO\_0008544$ ($Epidermis~development$) and $FMA\_67175$
($Anatomical~entity$)}
\label{fig:go_exp}
}
\end{figure*}

The logical difference also contains 80 new entailments that relate concepts
from domain ontologies (i.e. new cross-references). For example, the GO
concept $Epidermis~development$ is classified under the FMA concept
$Anatomical~entity$. This consequence is probably not intended and it is due to
the definition of range axioms in FMA (see Figure \ref{fig:go_exp}) and the use
of the property $part\_of$ in different scopes (in FMA relates anatomical entities, whereas in
GO biological processes).
Additionally, if a greater approximation of the logic difference is
considered (i.e. entailments of type \emph{(ii)}-\emph{(v)}) new consequences
are also obtained (e.g. $GO\_0030308 \sqsubseteq 
\dex{negatively\_regulates}GO\_0040007$, where GO\_0030308 stands for
\textit{Negative~regulation~of~cell~growth} and GO\_0040007 stands for
\textit{Growth}.
%
%

\section{Conclusions and future work}

The benefits of integrating phenotypic descriptions with domain ontologies 
have already discussed in the literature 
\cite{Lussier2004,MungallPheno2010,Hoehndorf2010}. However, the consequences
of the integration should be evaluated by domain experts in order to detect
potential unintended consequences.

In this paper we have performed a preliminary
evaluation\footnote{\hpomergedS and related domain ontology modules are
available at:
\url{http://krono.act.uji.es/people/Ernesto/phenotypeassessment/}} 
in which state of the art techniques (e.g. ontology reasoning, ontology
modularization, logical difference) have been reused to extract the set of new
consequences when integrating post-composed phenotypic descriptions, such as the
provided by \hpopc, with domain ontologies.
In a near
future, we intend to develop a system to guide the expert in the detection and
repair of unintended consequences such as in our previous tool
\textsf{ContentMap} \cite{ESWC2009}, in which we assessed the integration of
ontologies through mappings.

Moreover, domain ontologies contains cross-references (i.e. mappings)
which have not been considered for this preliminary assessment. These new
correspondences will probably lead to new consequences that
should be assessed. Thus, we also intend to adapt the techniques proposed in
\cite{umlsassessment10} to this new setting.



\bibliographystyle{splncs}
\bibliography{references}

\end{document}